\documentclass[letterpaper]{article} 
\usepackage{aaai25}  
\usepackage{times}  
\usepackage{helvet}  
\usepackage{courier}  
\usepackage[hyphens]{url}  
\usepackage{graphicx} 
\usepackage{natbib}  
\usepackage{caption} 
\usepackage{algorithm}
\usepackage{algorithmic}
\usepackage{amsfonts} 
\usepackage{multirow}
\usepackage[table,xcdraw]{xcolor}
\usepackage{booktabs}
\usepackage{subfigure}
\usepackage{bm}
\usepackage{tcolorbox}
\usepackage{arydshln}
\usepackage{amsmath}
\usepackage{tabularx}
\usepackage{makecell}
\usepackage{graphicx}
\newcommand{\std}[1]{}
\usepackage{newfloat}
\usepackage{listings}
\DeclareCaptionStyle{ruled}{labelfont=normalfont,labelsep=colon,strut=off} 
\floatstyle{ruled}
\newfloat{listing}{tb}{lst}{}
\floatname{listing}{Listing}
\title{DogLayout: Denoising Diffusion GAN for Discrete and Continuous Layout Generation}
\author{
    Zhaoxing Gan,
    Guangnan Ye,
}
\affiliations{
   School of Computer Science,Fudan University,Shanghai, China\\
   zxgan23@m.fudan.edu.cn
}

\usepackage{bibentry}

\begin{document}

\maketitle

\begin{abstract}
 Layout Generation aims to synthesize plausible arrangements from given elements. Currently, the predominant methods in layout generation are Generative Adversarial Networks (GANs) and diffusion models, each presenting its own set of challenges. GANs typically struggle with handling discrete data due to their requirement for differentiable generated samples and have historically circumvented the direct generation of discrete labels by treating them as fixed conditions. Conversely, diffusion-based models, despite achieving state-of-the-art performance across several metrics, require extensive sampling steps which lead to significant time costs. To address these limitations, we propose \textbf{DogLayout} (\textbf{D}en\textbf{o}ising Diffusion \textbf{G}AN \textbf{Layout} model), which integrates a diffusion process into GANs to enable the generation of discrete label data and significantly reduce diffusion's sampling time.  Experiments demonstrate that DogLayout considerably reduces sampling costs by up to 175 times and cuts overlap from 16.43 to 9.59 compared to existing diffusion models, while also surpassing GAN based and other layout methods. Code is available at https://github.com/deadsmither5/DogLayout.
\end{abstract}

%

\section{Introduction}
\label{sec:introduction}
Creating aesthetic layouts is an essential method for conveying important information to viewers and plays a critical role in various applications such as UI design \citep{rico}, advertisement poster synthesis \citep{poster}, and editing in printed media \citep{publaynet}. Traditionally reliant on manual design processes, the scalability and efficiency demands of modern digital content creation necessitate automated solutions using generative models. 

In the layout generation task, each layout element is characterized by discrete labels along with continuous size and position attributes. Besides conditioning on the label types, sizes, and positions, there are two other important layout generation tasks: 1) Unconditional generation, which involves generating layouts without predefined constraints or elements' attributes, such as randomly designing a game layout. 2) Completion, which involves filling in missing layout elements based on partially known elements and can be used to complete layouts forgotten by human designers.
\begin{figure}[htb]
    \centering
    \includegraphics[width=1\linewidth]{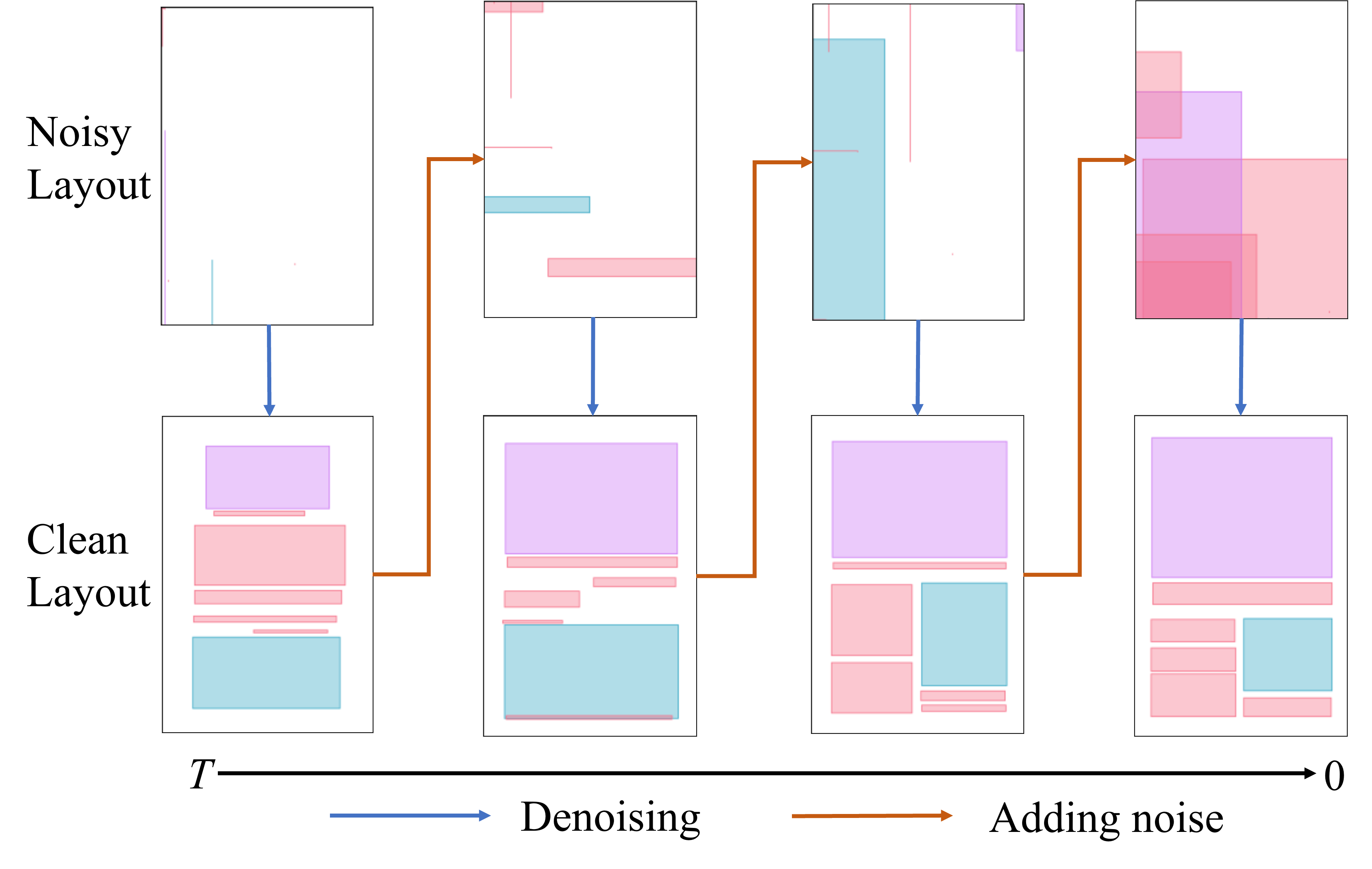}
\caption{Visualization of DogLayout's inference process. During inference, we first obtain the noisy layout from standard gaussian. Then the generator takes it as input to output the predicted clean layout. Subsequently, we derive the less noisy layout by adding noise to the predicted clean layout. Repeat the above process to achieve the final clean layout.}

    \label{fig:overall}
\end{figure}

In both unconditional generation and completion, generating the discrete label is inevitable. However, previous GAN-based layout models \citep{layoutgan,layoutgan++} can only generate the continuous sizes or positions conditioned on the label. Given a one-hot encoded label, GAN-based layout models fail to generate discrete data for two reasons: 1) Given the generator's probabilistic outputs, the discriminator's task is overly simplified as it only needs to identify a single non-zero element in the data's one-hot vector representation. 2) Using the non-differentiable argmax function to convert the generator's probabilistic outputs into discrete formats leads to vanishing gradients.

On the other hand, existing diffusion models \citep{layoutdm} focus too much on improving automatic evaluation indicators and ignore the importance of sampling speed in practical applications. To maintain the posterior distribution in a Gaussian form, diffusion models \citep{ddpm} always involve thousands of sampling timesteps. Although some works like LACE \citep{lace} use DDIM \citep{ddim} to accelerate the sampling process, the time cost is still significant. In scenarios where quick response is required, the high time cost is unaffordable. Additionally, the layouts generated by diffusion models, such as LayoutDM \citep{layoutdm}, still suffer from excessive overlap, which is highly noticeable to humans.

\begin{figure*}[t]
\centering
\includegraphics[width=1\textwidth]{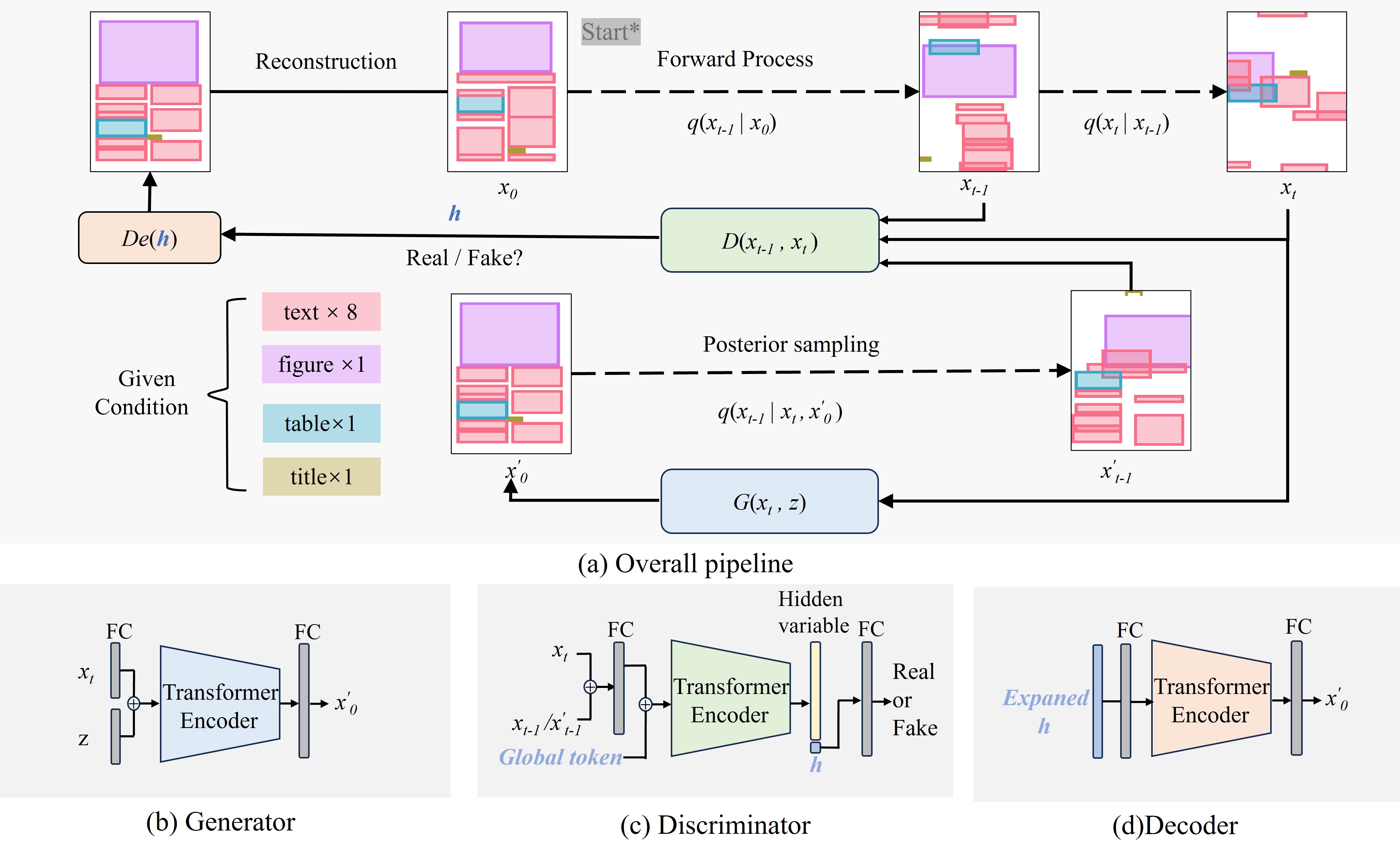} 
\caption{Overview of our method. (a) During training, we first obtain the noisy layout \(x_{t-1}\), then generate \(x_{t}\) by directly adding noise to \(x_{t-1}\). The generator then takes \(x_{t}\) and an additional latent dimension \(z\) as inputs to output the predicted clean layout \(x'_{0}\). Subsequently, we derive the predicted \(x'_{t-1}\) using \ref{eq:qxt-1xtx0}. For the real data, the discriminator evaluates the real noisy layout \(x_{t}\) and \(x_{t-1}\) to determine whether \(x_{t-1}\) is the true denoised layout of \(x_{t}\). An additional decoder then takes the global context token \(h\) from the discriminator and reconstructs \(x_{0}\), which forces the discriminator to learn the meaningful attributes of the layout. For the fake data, the discriminator assesses the real noisy layout \(x_{t}\) and the predicted layout \(x'_{t-1}\) to determine whether \(x'_{t-1}\) is the true denoised layout of \(x_{t}\). The model architectures are shown in (b), (c) and (d).}
\label{fig:pipeline}
\end{figure*}

In this study, we propose the DogLayout (shown in Figure~\ref{fig:pipeline}) to expand Layout GAN models' ability to handle unconditional generation and completion while maintaining a high sampling speed. By adding a diffusion process to GAN, we propose a new method for GANs to deal with discrete label data: 1) All operations on the Generator's output are differentiable. 
2) The discriminator does not directly see the generator's output and cannot distinguish real denoised layout from predicted denoised layout based solely on the presence of a single non-zero element.
Moreover, by using GAN to fit the non-Gaussian denoising distribution \citep{ddgan}, we significantly reduce the number of sampling steps and achieve a sampling speed that is up to 175 times faster than current diffusion-based layout models and also improved the overlap from 16.43 to 9.59.

We summarize our contributions as follows:
\begin{itemize}
    \item By adding a diffusion process to GANs, we propose a new method to generate discrete label data, which maintains the differentiability of the generator's output and prevents the discriminator from distinguishing real from fake data by detecting a single non-zero element.
    \item We expanded the capabilities of previous layout GAN models, which were limited to conditional tasks, to include unconditional and completion tasks, thereby improving the quality of generated layouts by up to 2.5 times compared to LayoutGAN++ \citep{layoutgan++}.
    \item Through extensive experiments, our model outperforms non-diffusion-based layout models in most tasks and reduces the time cost of layout generation by up to 175 times compared to diffusion-based layout models, while maintaining competitive performance. User studies indicate that our model is more favored by real users.
\end{itemize}

\section{Related Work}
\label{sec:related_work}
\subsection{Layout Generation}
Automatically generating layouts is a long-researched topic in graphic design \citep{automating_design}. Early studies optimize layouts by manually designing energy functions with constraints. Recent works have begun to use deep generative models to learn plausible layouts.  LayoutVAE \citep{layoutvae} introduces two conditional Variational Autoencoders (VAE). NDN-none \citep{ndn} is also a VAE-based model for conditional layout generation using graph neural networks. BLT \citep{blt}  proposes a hierarchical sampling policy with bidirectional layout transformer. As for the generative adversarial networks (GANs, \cite{gan}) based models, LayoutGAN \citep{layoutgan} can synthesize graphic layouts conditioned on different element attributes. LayoutGAN++ \citep{layoutgan++}  builds a transformer-based gan and formulates the layout generation as a constrained optimization problem. However traditional GAN can't deal with categorical data \citep{boundgan}, previous GAN based layout model is all limited to condition on the discrete labels. Recently, diffusion-based models have begun to be used. LayoutDM \cite{layoutdm}, LayoutDiffusion \citep{layoutdiffusion}  and \citep{unifylayoutdm} use the Discrete Diffusion Models \citep{d3pm} in a similar way to handle the structured layout data in the discrete representation, their works also expand the conditional layout generation to unconditional situation. LACE \citep{lace} proposed a novel align loss and also utilizes unconditional layout generation. Although diffusion-based layout models show strong capabilities in terms of diversity, the time cost and overlap in the generated layouts are non-negligible. Our work proposes a new way to enable GANs to generate discrete labels with minimal time cost.

\subsection{GANs for Discrete Data}
\paragraph{Challenges with Discrete Data in GANs.}Generative Adversarial Networks (GANs) face two main challenges when applied to discrete data. The generator transforms a latent vector into an output, while the discriminator evaluates whether this output resembles the actual discrete data, typically represented as a one-hot vector. The first challenge is that discriminator's task becomes straightforward—it only needs to detect the presence of a single non-zero element in the real one-hot discrete data. This simplicity contrasts sharply with generator's output, which often spreads non-zero probabilities across multiple dimensions. The second challenge arises when attempting to address the first: using argmax on generator's output to provide one-hot form input for discriminator leads to a gradient vanishing problem due to the non-differentiability of the argmax operation.

\paragraph{Existing solutions.}Several studies seek to address these challenges. Gumbel GAN\cite{gumbel_gan} employs the Gumbel reparameterization technique \citep{gumbel}, which enables gradient backpropagation from discriminator to generator. However, this approach introduces a new challenge: discriminator only observes the one-hot transformed output from generator, not the output itself, limiting it's ability to effectively guide generator's gradient optimization. BGAN \citep{boundgan} addresses this by exploring the boundaries of data distributions, yet accurately defining and identifying these boundaries remains challenging. Seq-GAN \citep{seqgan} uses discriminator as a reward function employing policy gradients, but designing an effective reward function that accurately evaluates the quality of generated layouts proves difficult. Figure~\ref{fig:discriminator_compare} shows that using LayoutGAN++ directly to model discrete label data results in gradient vanishing.

\section{Preliminary}
\paragraph{Problem Formulation.}
Following previous studies \citep{lace,layoutgan++}, we define a layout $l$ with $M$ elements and as $\{(c_{1},b_{1}),\dots,(c_{M},b_{M})\}$, where $(c_{i},b_{i})$ represents the $i$-th elements in $l$. $c_{i} \in \{0,\dots,N-1\}$ is the discrete label in a range of $N$ classes such as the Text or Title and $b_{i}=(x_{i},y_{i},w_{i},h_{i}) \in [0,1]^{4}$ is the corresponding center coordinates $(x, y)$ and size ratio $(w,h)$. 
\paragraph{Diffusion Models.}
A standard diffusion model contains a forward diffusion process and reverse diffusion process.
In diffusion's forward process, given $x_{0} \sim q(x_{0})$, each data $x_{t}$ is corrupted gradually by adding Gaussian noise to $x_{t-1}$. In reverse process , the goal is to  train a network $p_{\theta}$ to predict $x_{t-1}$ from $x_{t}$ and the whole training objective could be presented by minimizing the Evidence Lower Bound:
\begin{gather}
    \mathcal{L}_\text{{ELBO}} = \underbrace{\mathbb{E}_{q}[D_\text{{KL}}(q(x_{T}|x_{0})||p_{\theta}(x_{T}))]}_{\mathcal{L}_{T}} - \underbrace{\log p_{\theta}(x_{0}|x_{1})}_{\mathcal{L}_{0}} +\notag \\ \underbrace{\mathbb{E}_{q}[\sum_{t=2}^{T}D_\text{{KL}}(q(x_{t-1}|x_{t})||p_{\theta}(x_{t-1}|x_{t}))]}_{\mathcal{L}_{t-1}}.
    \label{eq:elbo}
\end{gather}

In a standard diffusion process, when forward timestep is large enough $q(x_{T}|x_{0})$ and $p_{\theta}(x_{T})$ will both follows a standard normal distribution $\mathcal{N}(0,1)$ and the $\mathcal{L}_{T}$ term will be nearly equal to zero. $\mathcal{L}_{0}$ is the reconstruct loss from $x_{1}$ to $x_{0}$.  $\mathcal{L}_{t-1}$ means to train a neural network $p_{\theta}(x_{t-1}|x_{t})$ to fit the true distribution $q(x_{t-1}|x_{t})$.  In order to trace  $q(x_{t-1}|x_{t})$ , DDPM \citep{ddpm}  use $q(x_{t-1}|x_{t},x_{0})$  to alternate $q(x_{t-1}|x_{t})$, which could be denoted as:
\begin{equation}
    q(x_{t-1}|x_{t},x_{0}) = \frac{q(x_{t}|x_{t-1},x_{0})q(x_{t-1}|x_{0})}{q(x_{t}|x_{0})}
    \label{eq:qxt-1xtx0}.
\end{equation}

\section{Approach}
\label{sec:DogLayout}

 \subsection{Overview and Model Architecture}
DogLayout builds on Diffusion GAN models \citep{ddgan,semi_gan}. In this chapter, we will first introduce the model architecture of DogLayout, then discuss the details of our framework and the methods used to reduce sampling time costs by integrating a GAN into the diffusion process, and explain how this integration enables GANs to handle discrete data. 
 \paragraph{Conditional and Unconditional Generation.}
Conditional generation involves creating an entire layout from a partially known layout \(x_p\). Let \(m\) represent the mask, where $1$ and $0$ indicate known and unknown layout attributes, respectively. The conditional information is incorporated as follows: \(x_{t-1} = (1-m) \odot \tilde{x}_{t-1} + m \odot x_p\), where \(\tilde{x}_{t-1} \sim p_{\theta}(x_{t-1} | x_t)\).
Unconditional generation refers to the process of generating a layout initially from standard Gaussian.
\paragraph{Generator.}
To process the input noise layout $x_t$, we utilize a fully-connected layer to expand its dimensions to the embedding dimension. The latent variable \(z\) is initially sampled from a standard Gaussian distribution, subsequently resized to the specified latent dimension through a fully-connected layer. while temporal embedding is not explicitly incorporated. The core processing unit comprises a transformer-encoder \citep{attention} Finally, the transformer-encoder's output is adjusted back to the input's dimensions using another fully-connected layer:
\begin{gather}
    z\sim \mathcal{N}(0, I), \; h_{1} = f_{FC}(z), \notag\\
    h_{2} = f_{FC}(x_{t}), \notag\\
    h_{3} = f_{TF-ENC}([h_{1},h_{2}]), \notag\\
    x'_{0} = f_{FC}(h_{3}). 
\end{gather}
In the above notations, $f_{FC}$ represents the fully-connected layer, and $f_{TF-ENC}$ represents the transformer-encoder layer. $x_{t}$ is sampled from $ \mathcal{N}(x_{t};\sqrt{1-\beta_{t}}x_{t-1},\beta_{t} \textbf{I})$ and condition is injected from known layout \(x_p\).
\paragraph{Discriminator.}
The discriminator's input is formed by concatenating \(x_t\) with either \(x_{t-1}\) or \(x^{'}_{t-1}\), depending on whether the data is real or generated. This combined input is then passed through a fully-connected layer to expand its dimensions to match the embedding dimension. Position embedding is injected via a trainable embedding layer, while time embedding is not included. The core unit consists a transformer-encoder which includes a learnable special token $h_{s}$ to get global context token \(h\). Then a fully-connected layer processes \(h\) to produce the probability logits:
\begin{gather}
    h_{1} =f_{FC}([x_{t-1},x_{t}])\; or \; f_{FC}([x^{'}_{t-1},x_{t}]), \notag \\
    [h,h_{2}] = f_{TF-ENC}(h_{1},h_{s}), \notag \\
    p = f_{FC}(h).
\end{gather}
Here, p presents the probability that whether \(x_{t-1}\) or \(x^{'}_{t-1}\) is the true denoised layout of \(x_{t}\).
\paragraph{Decoder.}
When the discriminator processes real inputs \(x_t\) and \(x_{t-1}\), it employs a transformer-encoder and a fully-connected layer to reconstruct the initial layout \(x_0\) from the extracted global context $h$:
\begin{gather}
    h_{1} = f_{TF-ENC}(h), \notag \\
    x'_{d0} = f_{FC}(h_{1}).
\end{gather}
$ x'_{d0}$ is the reconstruction results of decoder. This reconstruction process enables the discriminator to learn the meaningful attributes of the layout effectively, which enables the discriminator to effectively distinguish between real and generated layouts based on their meaningful attributes.

\subsection{DogLayout}
\label{subsec:doglayout}
The key to reducing sampling time in the diffusion process is to decrease the timesteps. Using Bayes' rule, the real denoising distribution \(q(x_{t-1}|x_{t}) = q(x_{t}|x_{t-1})q(x_{t-1})/q(x_{t})\), when \(T\) is sufficiently large, the noise added between each adjacent step is small enough that the ratio \(q(x_{t-1})/q(x_{t}) \approx 1\). Consequently, both \(q(x_{t}|x_{t-1})\) and \(q(x_{t-1}|x_{t})\) can be assumed to follow Gaussian distributions.\

To reduce the timestep \(T\) to a smaller number (e.g., \(T = 4\)), we can use a GAN to match the non-Gaussian distribution \(q(x_{t-1}|x_{t})\). When \(T\) is small, DDGAN \citep{ddgan} proposes using a conditional generative adversarial network to minimize the distance between these two distributions instead of the original KL Divergence described in Equation~\ref{eq:elbo}. Given the noisy layout \(x_{t}\) to both the generator and discriminator, the generator \(p_{\theta}(x_{t-1}|x_{t})\) aims to reconstruct the cleaner layout \(x_{t-1}\) that is indistinguishable from the real \(x_{t-1}\). The discriminator aims to maximize its ability to distinguish between the real cleaner layout \(x_{t-1}\) and the predicted \(x_{t-1} \sim p_{\theta}(x_{t-1}|x_{t})\). This training process can be regarded as minimizing the following expression, where \(D_{adv}\) represents a metric for calculating the distance between two distributions (e.g., Wasserstein distance \citep{wgan}):
\begin{equation}
    \underset{\theta}{\min}\sum_{t\geq 1} \mathbb{E}_{q(x_{t})}[D_{adv}(q(x_{t-1}|x_{t}), p_{\theta}(x_{t-1}|x_{t}))].
\end{equation}

We choose the softened reverse KL as the \(D_{adv}\). We propose not to inject time into the generator and discriminator due to the fact that the timestep \(t\) is implicitly included in the noise strength of the given \(x_{t}\). The generator will take an additional \(N\)-dimensional latent variable \(z\) to enhance diversity and directly output the predicted version of the layout \(x_{0} = G_\theta(x_{t}, z)\). Then, \(x_{t-1}\) is sampled using Equation~\ref{eq:qxt-1xtx0}. The denoising distribution \(p_\theta(x_{t-1} | x_t)\) can be written as:
\begin{gather}
p_\theta(x_{t-1} | x_t) := \int p_\theta(x_0 | x_t) q(x_{t-1} | x_t, x_0) \, dx_0 = \notag \\
\int p(z) q(x_{t-1} | x_t, x_0 = G_\theta(x_t, z)) \, dz.
\end{gather}

Inspired by the self-supervised learning method of \citep{selfsupervisedgan}, when trained with real \(x_{t-1}\) and \(x_{t}\), another decoder takes the global context token \(h\) from the discriminator and reconstructs the layout \(x_{0} = De(h)\). With such a constraint, we can ensure that the discriminator has learned effective layout features. The training objective for the discriminator is described as:
\begin{gather}
    \min\sum_{t\geq1}\mathbb{E}_{q(x_{t})}[\mathbb{E}_{p_{\theta}(x_{t-1}|x_{t})}[-\log(1-D(x_{t-1},x_{t}))] + \notag \\ 
    \mathbb{E}_{q(x_{t-1}|x_{t})}[-\log(D(x_{t-1},x_{t}))] + \notag \\
    \mathbb{E}_{q(x_{0}|x_{t})}[\mathcal{L}_{rec}(x_{0},De(h))]]. 
    \label{eq:dis_loss}
\end{gather}

\paragraph{DogLayout for Discrete Data.}
 We are the first to discover that adding a diffusion process to GANs enables the generation of discrete data. The introduction of the diffusion process addresses two challenges that GANs face with discrete data, for two specific reasons:
\begin{enumerate}
\item All operations on the generator's output \(x_{0} = G(x_{t}, z)\) are differentiable. Instead of applying an argmax to \(x_{0}\), we use Equation~\ref{eq:qxt-1xtx0} to compute the predicted noisy layout \(x_{t-1}\). Meanwhile, the operations of the discriminator are all on \(x_{t-1}\), ensuring that the gradient flows normally towards the generator after back-propagation.
\item The discriminator no longer directly sees the output of the generator, except when \(T = 1\). Since all noisy layouts \(x_{t-1}\) are defined in a continuous space, the discriminator cannot distinguish real noisy \(x_{t-1}\) from predicted noisy \(x_{t-1}\) by simply detecting whether they contain a single non-zero element.
\end{enumerate}
Technically, we can now directly treat the one-hot label \(l_0\) in whole layout \(x_0\) as a class prototype. After directly applying \(T\) step noise to \(l_0\), we can use the generator \(p_{\theta}(l_{t-1} | l_{t})\) to gradually reconstruct \(l_0\) from \(l_T\). The final discrete layout label \(L\) is sampled by the following equation:

\begin{gather}
    L = \arg\max_i (p(l = i | l_0)), \notag \\
    \quad \text{where } p(l = i | l_0) = \frac{\exp(l_0[i])}{\sum_{j=1}^{K} \exp(l_0[j])}.
\end{gather}

\paragraph{Variable Length Generation.}
Diffusion GAN models are designed to produce data with fixed dimensions; therefore, they do not naturally extend to layout generation, where the number of elements in each layout can vary. We add an additional padding element with an extra noise class \(c = N\) and \(b = (0, 0, 0, 0)\) to manage variable length generation for up to a maximum of \(E\) elements.

 \label{sec:implementation}

\section{Experiments}
\label{sec:experiments}

\subsection{Experimental Setup}

\paragraph{Datasets.}To evaluate our method, We apply two widely-used datasets in recent studies about layout generation. \textbf{Rico} \citep{rico} contains 72K mobile app screens collected from over 97K Android apps with 25 label types such as Advertisement, Text Button and so on. \textbf{PubLayNet} \citep{publaynet} contains 330K document layout annotations sourced from PubMed Central with 5 label types such as table, figure and list. 
We set the maximum number of elements to 25. The dataset split of training, validation and test is 35, 851/2, 109/4, 218 for Rico and 315, 757/16, 619/11, 142 for PubLayNet.

\paragraph{Evaluation Metrics.}
We adapt the five metrics to measure the quality of the generated layouts. These include: 
\textbf{Fréchet Inception Distance (FID)} \citep{fid}, which measures the distance between feature vectors calculated for real and generated layouts; for this, we use the same feature extraction model employed in LayoutDM \citep{layoutdm}.
\textbf{Alignment}, which evaluates the aesthetic and functional arrangement of elements within the layout. 
\textbf{Maximum Intersection over Union (IoU)}, which quantifies the overlap accuracy between the predicted and actual bounding boxes of layout elements. 
\textbf{Overlap}, which quantifies the intersection area among elements within a layout. However, given that overlapping elements are a prevalent pattern within the Rico dataset, this metric is not appropriate for evaluating performance on Rico. Therefore, it is exclusively employed to assess generation quality on the PubLayNet dataset. 
\textbf{Generation Time per Sample (T/sample)}, which compares the time cost per sample across models.

\paragraph{Tasks.}
Our research explores multiple tasks in layout generation: 
 1) Conditional Generation by Class \textbf{(C→S+P)}: Given a class of elements, generates size and position. 
 2) Conditional Generation by Class and Size \textbf{(C+S→P)}: Conditions on both the class and size of elements to generate positions. 
 3) \textbf{Completion} with Attributes for a Subset of Elements: Employs a binary condition mask, as described in LayoutDM \citep{layoutdm}, by randomly selecting 0\% to 20\% of elements from real samples and generate entire layout based on the masked layout.
 4) Unconditional Generation \textbf{(Uncond)}: Generates layouts without any conditions.

\paragraph{Baselines.}We compare DogLayout against a suite of models on layout tasks using the PubLayNet and Rico datasets, categorized into: 
\textbf{Task-Specific models} include LayoutVAE \citep{layoutvae}, NDN-none \citep{ndn}, and LayoutGAN++ \citep{layoutgan++}, which could prove DogLayout's improvements on GAN-based model. 
\textbf{Task-Agnostic models} comprise BLT \citep{blt}, MaskGIT \citep{maskgit} and LayoutDM \citep{layoutdm}, which is the state-of-art diffusion-based model.

\paragraph{Implementation Details.} We set the transformer-encoder in generator and decoder with 4 layers, 8 attention heads, and a feed-forward network width of 2048, while 8 layers, 4 heads for discriminator. Embedding dims for label and bounding box are both 256. We find $T = 4$ is enough for the c→s+p and c+s→p task. But generator needs more timesteps to match the distribution of discrete label. Thus the timestep for unconditional and completion task is $T=8$ for PubLayNet and $T=12$ for Rico. We train our model on a single NVIDIA RTX 4090 with 24 GB memory and Adam optimizer with learning rate $10^{-5}$. The batch size is 512. The model on each task is trained for up to 200 epochs.

\subsection{Quantitative Comparisons}
We report the overlap results on the PubLayNet dataset in Table~\ref{tab:overlap}. Our method demonstrates a significant improvement in the overlap metric compared to LayoutGAN++ and LayoutDM. It is crucial to emphasize the significance of the overlap metric, as overlaps are more noticeable to humans than FID scores and can significantly affect the aesthetics and comfort of the layout. The layouts generated by our model exhibit considerably less overlap, which is particularly important in the context of document layout design.
\setlength\tabcolsep{3pt}
\begin{table}[h]
  \small
  \centering
  \begin{tabular}{lccccc}
    \toprule
    \multicolumn{1}{r}{Task} & C → S+P & C+S → P & Completion & Uncond \\
    \cmidrule(lr){2-5} 
    \multicolumn{1}{r}{Model\hspace{4mm} Metric} & Overlap & Overlap & Overlap & Overlap \\   
    \midrule  
    LayoutGAN ++   & 22.8 & 17.6 & - & - \\

    LayoutDM     &  16.43 & 18.91 &15.0 & \textbf{13.43}\\

    DogLayout(Ours)       &\textbf{9.59}& \textbf{12.5} &\textbf{11.9} &16.3                                         \\ 
    \bottomrule
  \end{tabular}\\
   \caption{Overlap comparison of our model with LayoutGAN++ and LayoutDM on PubLayNet dataset.}
   \label{tab:overlap}
\end{table}
\setlength\tabcolsep{6pt}
\setlength{\tabcolsep}{1.5pt}
\begin{table*}[htbp]
\centering
\vspace*{-2mm}
\footnotesize
\begin{tabular}{lcccccccccccccccc} \toprule
\multicolumn{1}{r}{Task} & \multicolumn{4}{c}{C $\rightarrow$ S+P} & \multicolumn{4}{c}{C+S $\rightarrow$ P} & \multicolumn{4}{c}{Completion} & \multicolumn{4}{c}{Uncond} \\
\cmidrule(lr){2-5} \cmidrule(lr){6-9} \cmidrule(lr){10-13} \cmidrule(lr){14-17}
\multicolumn{1}{r}{Dataset} & \multicolumn{2}{c}{Rico} & \multicolumn{2}{c}{PubLayNet} & \multicolumn{2}{c}{Rico} & \multicolumn{2}{c}{PubLayNet} & \multicolumn{2}{c}{Rico} & \multicolumn{2}{c}{PubLayNet}  & \multicolumn{2}{c}{Rico} & \multicolumn{2}{c}{PubLayNet} \\
Model
& FID $\downarrow$ & Max. $\uparrow$ & FID $\downarrow$ & Max. $\uparrow$
& FID $\downarrow$ & Max. $\uparrow$ & FID $\downarrow$ & Max. $\uparrow$
& FID $\downarrow$ & Max. $\uparrow$ & FID $\downarrow$ & Max. $\uparrow$
& FID $\downarrow$ & Align. $\downarrow$ & FID $\downarrow$ & Align. $\downarrow$\\
\midrule

LayoutVAE
& 33.3 & 0.249 & 26.0 & 0.316
& 30.6 & 0.283 & 27.5 & 0.315
& - & - & - & -
& - & - & - & - \\

NDN-none
& 28.4 & 0.158 & 61.1& 0.162
& 62.8 & 0.219 & 69.4& 0.222
& - & - & - & -
& - & - & - & - \\

LayoutGAN++
& 6.84 & 0.267 & 24.0 & 0.263
& 6.22 & 0.348 & 9.94 & 0.342
& - & - & - & -
& - & - & - & -\\

MaskGIT
& 26.1 & 0.262 & 17.2 & \textbf{0.319}
& 8.05 & 0.320 & 5.86 & 0.380
& 33.5 & 0.533 & 19.7 & \textbf{0.484}
& 52.1 & \textbf{0.015} & 27.1 & \textbf{0.101}
\\

BLT
& 17.4 & 0.202 & 72.1 & 0.215
& 4.48 & 0.340 & 5.10& \textbf{0.387}
& 117 & 0.471 & 131 & 0.345
& 88.2 & 1.030 & 116 & 0.153
\\

LayoutDM
& 4.63 & \textbf{0.274} & \textbf{8.96} & 0.308
& \textbf{2.26} & \textbf{0.390} & \textbf{4.29} & 0.379
& 10.5 & \textbf{0.576} & \textbf{8.47} & 0.376
& \textbf{8.96} & 0.162 & 15.5 & 0.195
\\
DogLayout(Ours) 
&  \textbf{4.44} & 0.268 & 9.62 & 0.287
& 2.86 & 0.362 & 9.47 & 0.345 
& \textbf{10.3} & 0.473 & 16.8 & 0.356
& 10.9 & 0.212 & \textbf{14.7} & 0.190
\\

\midrule

Real data & 1.85 & 0.691 & 6.25 & 0.438 & 1.85 & 0.691 & 6.25 & 0.438 & 1.85 & 0.691 & 6.25 & 0.438 &1.85 &0.109 &6.25 &0.021\\

\bottomrule \end{tabular}
\caption{
      Quantitative results on the PubLayNet and Rico datasets for four generation tasks. Task-specific models include LayoutVAE, NDN-none, and LayoutGAN++, which must be conditioned on class labels. The best results are highlighted in \textbf{bold}.
}
\label{tab:quantitative_results}
\end{table*}
\setlength\tabcolsep{6pt}
\\ \noindent We summarize other quantitative comparison results in Table~\ref{tab:quantitative_results}. Compared to LayoutGAN++, our model not only expands GAN's capabilities for completion and unconditional tasks but also significantly improves the FID and MaxIoU scores.  Specifically, in the C→S+P task on PubLayNet, we reduced LayoutGAN++'s FID score from 24.0 to 9.62, achieving nearly a 2.5 times improvement. Our model outperforms all non-Diffusion models, with a few exceptions in tasks involving MaskGIT and BLT. Regarding diffusion models, our model exhibits comparable performance . DogLayout also outperforms LayoutDM in several FID metrics and demonstrates significant improvements in sampling speed, as detailed in Table~\ref{tab:sampling_time}. Additionally, we observed suboptimal performance in the completion task on PubLayNet, which we suspect is due to the larger continuous space compared to LayoutDM. 

\paragraph{Sampling Time Comparison.}
In Table~\ref{tab:sampling_time}, we illustrate the sampling time efficiency of our DogLayout compared to LayoutGAN++ and LayoutDM. The experiments were conducted over 50 epochs with a batch size of 128. Our results indicate that DogLayout significantly outperforms the diffusion models, being 175 times faster than LayoutDM in conditional generation tasks. Additionally, it demonstrates substantial speed improvements in other tasks as well. Compared to LayoutGAN++, our model incurs only a minimal increase in sampling time while extending capabilities to include unconditional and completion tasks, thereby significantly enhancing performance in conditional tasks. This improvement in efficiency across tasks positions our model as particularly suitable for real-world applications, where rapid response times are crucial and can significantly conserve resources while maintaining high output quality.

\setlength\tabcolsep{2pt}
\begin{table}[]
  \small
  \centering
  \begin{tabular}{lccccc}
    \toprule
    Model  & LayoutGAN++ & LayoutDM  & \multicolumn{3}{c}{DogLayout(Ours)}\\   
    \midrule  
    Timesteps     & 1 & 50 & 4 & 8 & 12 \\
    
    T/sample (ms)     & 0.0327 & 23.3 & 0.133 & 0.255 & 0.377 \\
    \bottomrule
  \end{tabular}\\
  \caption{Sampling time comparison of our model with LayoutGAN++ and LayoutDM. ($T = 4$  for C→S+P and C+S→P on both dataset. $T = 8/12$ for uncond and completion on PubLayNet/Rico respectively.)}
  \label{tab:sampling_time}
\end{table}
\setlength\tabcolsep{6pt}
\begin{figure}
    \centering
    \includegraphics[width=1\linewidth]{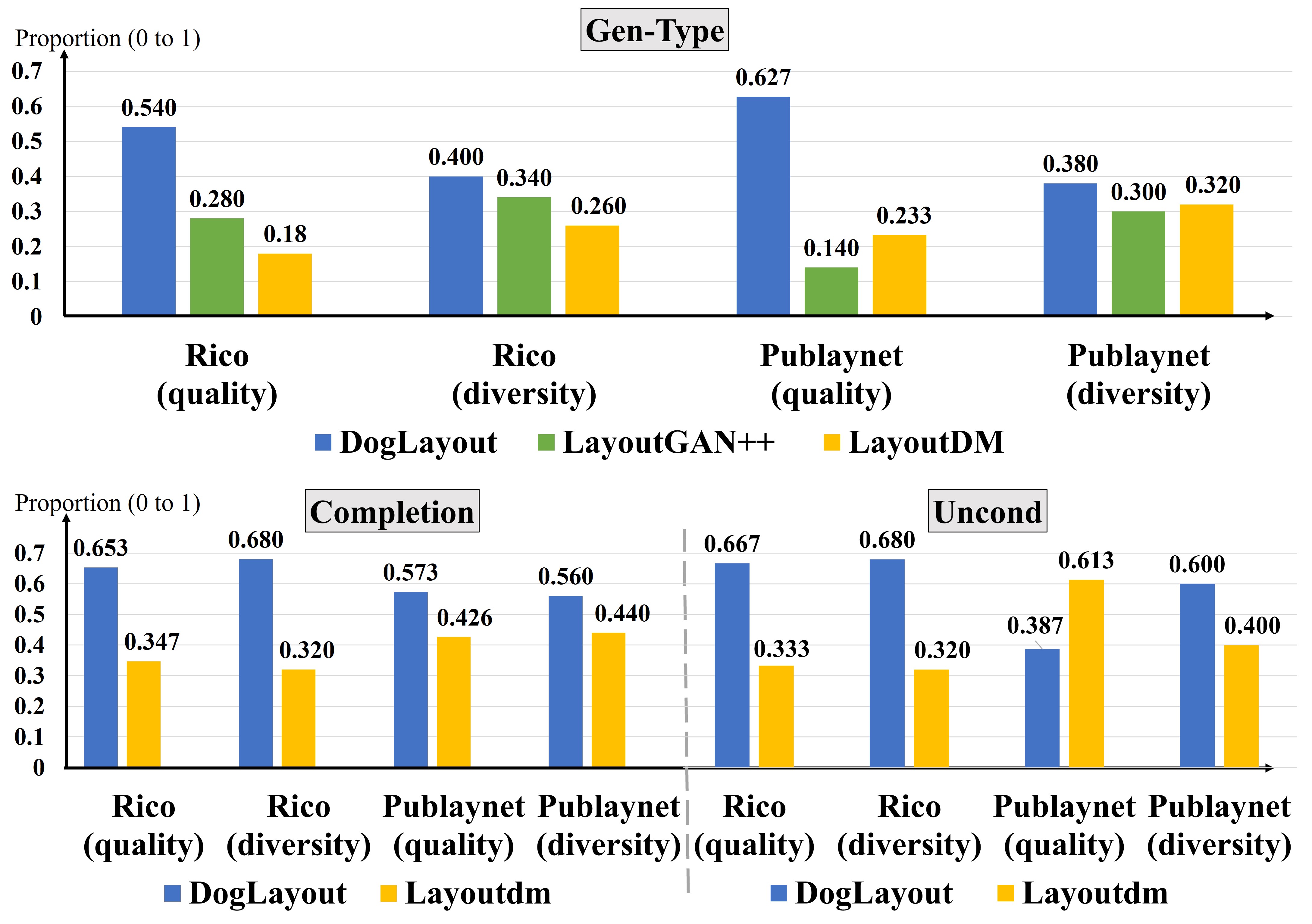}
    \caption{Results of the user studies for Gen-Type(C→S+P and C+S→P), Completion and Uncond. For every selection, We place samples from multiple models and count how many people prefer the layouts generated from each model. }
    \label{fig:userstudy}
\end{figure}
\subsection{Qualitative Comparisons}
\begin{figure*}[ht]
    \centering
    \includegraphics[width=1\linewidth]{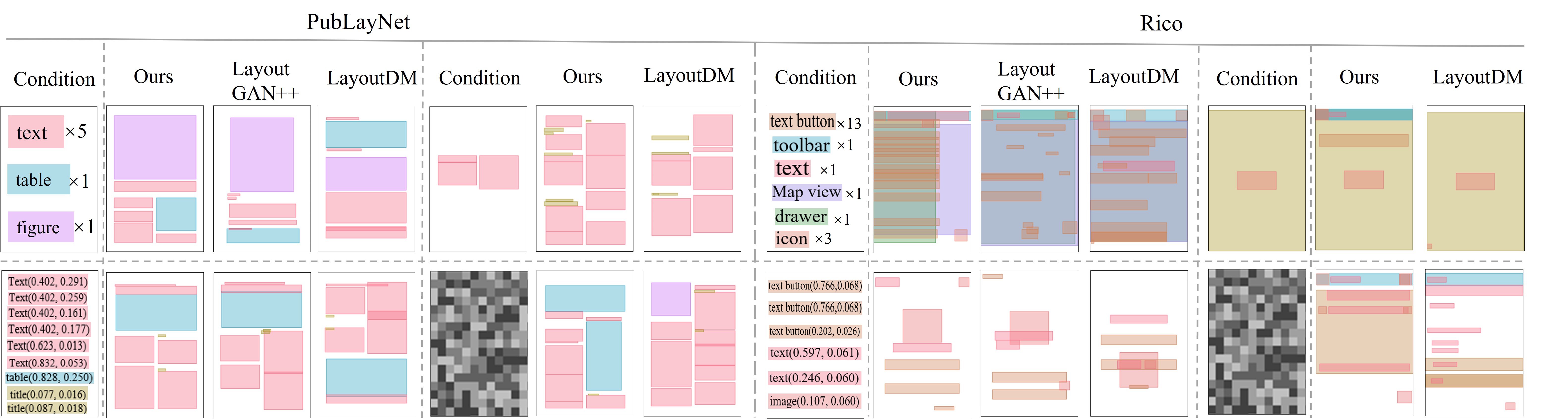}
    \caption{Qualitative comparison results on Rico and PubLyNet for four generation tasks with LayoutGAN++ and LayoutDM. Different colors represent different label classes and the decimals in parentheses are the values of width (w) and height (h).}
    \label{fig:qualitative_result}
\end{figure*}

We also present a qualitative comparison of our model with LayoutGAN++ and LayoutDM in Figure~\ref{fig:qualitative_result}. Compared to LayoutGAN++, our model produces layouts with superior spatial arrangement. It can also generate aesthetic layouts for completion and unconditional tasks, a feat not achievable by LayoutGAN++. Notably, while LayoutDM produces layouts with significant overlap areas on PubLayNet, DogLayout generates layouts with much less overlap, enhancing their visual appeal and harmony. We hypothesize that this is because the discriminator can easily detect instances of layout overlap and guide the generator to improve.

\paragraph{User Study.} We conducted a user study, as shown in Figure~\ref{fig:userstudy}, aimed at assessing the aesthetics and diversity of generated layouts, attributes that are often not fully captured by automated metrics such as FID. We designed two types of evaluations: quality and diversity, and invited 25 participants. For each task, samples were randomly selected from each model's outputs generated in the same batch. For the quality test, participants were asked to select the most aesthetically pleasing layout from a set of anonymized layouts. For the diversity assessment, participants were shown 20 layouts generated in the same batch and asked to identify the set with the greatest diversity. Results suggest that DogLayout is more favored by real users.
\subsection{Ablation Study}
\begin{figure}
    \centering
    \includegraphics[width=1\linewidth]{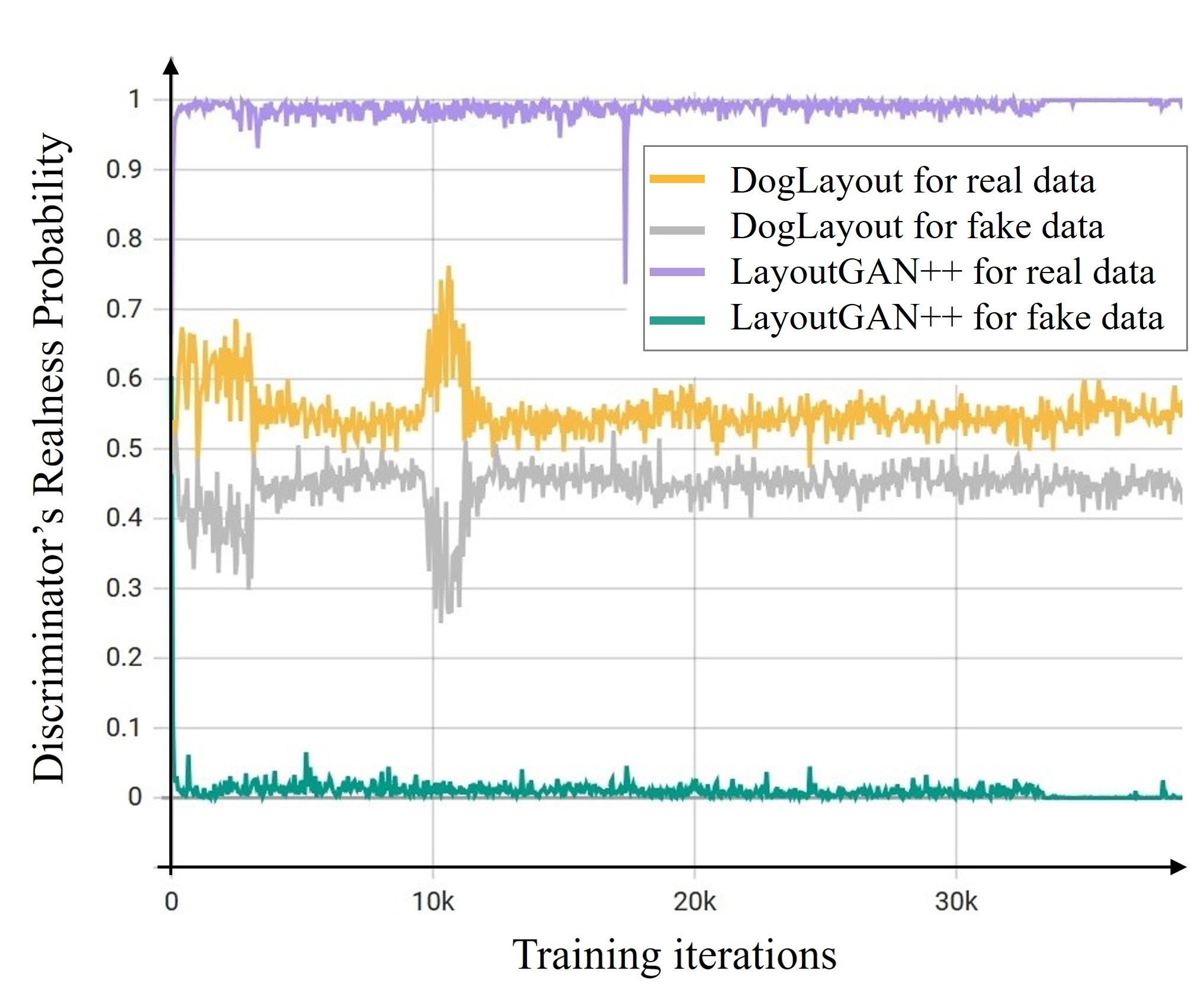}
    \caption{ Comparative analysis of discriminator’s ability to distinguish real and fake/generated data during training for LayoutGAN++ and DogLayout.}
    \label{fig:discriminator_compare}
\end{figure}

\paragraph{Discrete Layout Generation.}
Figure~\ref{fig:discriminator_compare} illustrates the discriminator’s probability of recognizing real and fake data during the training process. In the absence of a diffusion process for discrete data, LayoutGAN++ fails to generate discrete labels. Initially, the discriminator of LayoutGAN++ swiftly approaches a realness probability close to $1$ for real data and $0$ for fake data, indicating that the gradients quickly vanish. In contrast, DogLayout exhibits remarkable training stability, ensuring consistent convergence and performance improvements across epochs.

\paragraph{Number of Denoising Timesteps.}
As illustrated in Table~\ref{tab:ablation_time}, the FID score significantly improves with increasing timesteps, reaching a minimum at \(T=8\). Beyond \(T=8\), the FID score slightly increases, We hypothesize that when \( t \) is too large, both \( x_{t} \) and \( x_{t-1} \) are close to pure noise, making it difficult for the decoder to reconstruct \( x'_{0} \). Consequently, this leads to challenges for the discriminator in learning meaningful layout attributes. When \(T=2\), the alignment is low because most layout samples are blank. When \(T\) is too low, DogLayout is unable to fit the non-Gaussian distribution in unconditional generation. These results underscore the importance of selecting an optimal number of timesteps to balance generation quality and computational efficiency.

\begin{table}
\centering 
\begin{tabular}{lcc}
    \toprule
    Timestep  & FID↓  & Align↓  \\
    \midrule
    T = 2    &199 &  0.0619       \\
    T = 4    &53.3 &  0.450  \\
    T = 6    &37.4 &  0.207  \\
    T = 8    & \textbf{14.7} &  0.190    \\
    T = 10   &19.9 &  \textbf{0.179}    \\
    \bottomrule
\end{tabular}
\captionof{table}{Timestep effect on PubLayNet dataset with unconditional generation.}
\label{tab:ablation_time}
\end{table}

\section{Conclusion}
\label{sec:conclusion}
In this work, we propose DogLayout and introduce a new method for GANs to generate discrete label data. We expand the capabilities of previous conditional-only layout GAN models to unconditional and completion tasks, while significantly reduce the diffusion sampling timesteps. Our model achieves speeds up to 175 times faster and better overlap than current diffusion models while maintaining high quality and diversity. We foresee wide-ranging applications of our discrete handling method for GANs, potentially extending beyond the layout domain into other areas where discrete data is prevalent, such as structured data synthesis.
\section{Limitation}
While our model has several advancements compared to previous works, there are still several limitations. First, there is still room for improvement in automatic evaluation metrics. Second, most existing layout works lack content awareness, which means the generated layout elements have limited interaction with the background image. In future work, we will expand our model's ability and add image-layout cross attention to improve the current model's content awareness. 
\appendix

\bigskip

\bibliography{aaai25} 

\end{document}